\documentclass[10pt,twocolumn,letterpaper]{article}

\usepackage{cvpr}              

\usepackage{graphicx}
\usepackage{amsmath}
\usepackage{amssymb}
\usepackage{booktabs}
\usepackage{verbatim}
\usepackage{multirow}
\usepackage{color}
\usepackage{wrapfig}
\usepackage{makecell}

\usepackage{color, colortbl}
\definecolor{LightCyan}{rgb}{0.88,1,1}

\usepackage[pagebackref,breaklinks,colorlinks]{hyperref}

\usepackage[capitalize]{cleveref}
\crefname{section}{Sec.}{Secs.}
\Crefname{section}{Section}{Sections}
\Crefname{table}{Table}{Tables}
\crefname{table}{Tab.}{Tabs.}

\def\cvprPaperID{**} 
\def\confName{CVPR}
\def\confYear{2022}

\newcommand{\myparagraph}[1]{\vspace{3pt}\noindent{\bf #1}}

\begin{document}

\title{A Unified Query-based Paradigm for Point Cloud Understanding}

\author{Zetong Yang$^{1\ast}$~~~~~
Li Jiang$^{2\ast}$~~~~~
Yanan Sun$^{3}$~~~~~
Bernt Schiele$^{2}$~~~~~
Jiaya Jia$^{1}$
\\
$^{1}$CUHK~~~~~~~~~$^{2}$MPI Informatics~~~~~~~~~$^{3}$HKUST\\
\vspace{-2mm}
{\tt\small 
\{tomztyang, now.syn\}@gmail.com~
\{lijiang, schiele\}@mpi-inf.mpg.de~
leojia@cse.cuhk.edu.hk
} 
}

\maketitle

\begin{abstract}
3D point cloud understanding is an important component in autonomous driving and robotics. In this paper, we present a novel \textbf{E}mbedding-\textbf{Q}uerying paradigm (EQ-Paradigm) for 3D understanding tasks including detection, segmentation and classification. EQ-Paradigm is a unified paradigm that enables the combination of any existing 3D backbone architectures with different task heads. Under the EQ-Paradigm, the input is firstly encoded in the embedding stage with an arbitrary feature extraction architecture, which is independent of tasks and heads. Then, the querying stage enables the encoded features to be applicable for diverse task heads. This is achieved by introducing an intermediate representation, i.e., Q-representation, in the querying stage to serve as a bridge between the embedding stage and task heads. We design a novel Q-Net as the querying stage network. Extensive experimental results on various 3D tasks show that EQ-Paradigm in tandem with Q-Net is a general and effective pipeline, which enables a flexible collaboration of backbones and heads, and further boosts the performance of the state-of-the-art methods. Codes and models are available at \href{https://github.com/dvlab-research/DeepVision3D}{https://github.com/dvlab-research/DeepVision3D}.

{\let\thefootnote\relax\footnotetext{$^\ast$ Equal contribution.}}

\begin{figure}[t]
  \centering
  \includegraphics[width=1.0\linewidth]{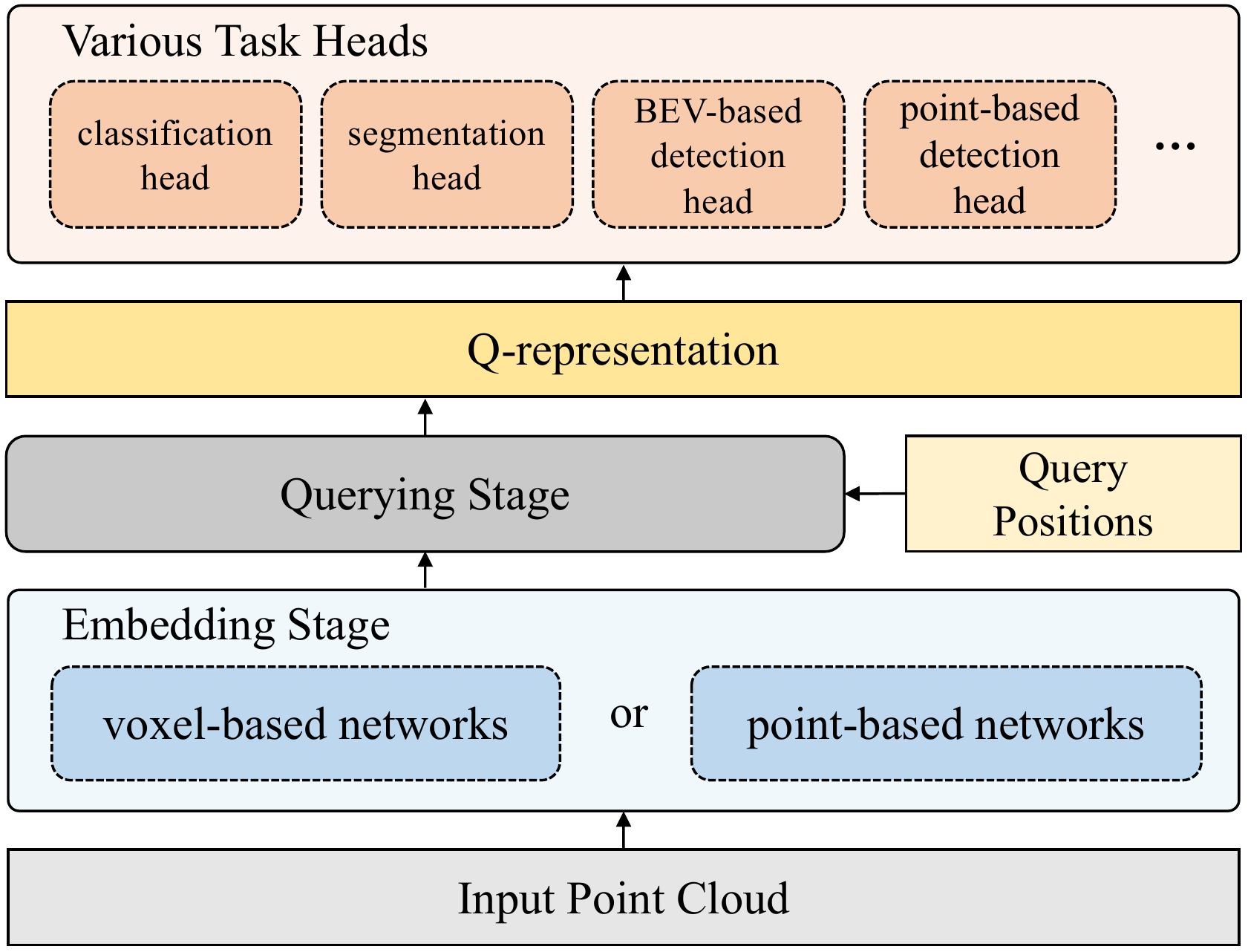}\\
  \vspace{-0.1in}
  \caption{Illustration of the unified query-based EQ-Paradigm. The query position can be randomly designated in the 3D scene, thus making it possible to combine any backbone embedding networks with different task heads. }
  \label{fig:teaser}
  \vspace{-0.2in}
\end{figure}
\end{abstract}

\section{Introduction}
3D point cloud understanding tasks have attracted much attention recently since they benefit many real-life applications like autonomous driving \cite{KITTIDATASET1}, robotics \cite{GraspNet} and augmented reality \cite{Multiple3Dtracking}.
In point cloud understanding, there are two dominant input representations: points and voxels. %
Specifically designed for these two representations, mainstream models can be grouped into two types:  point-~\cite{POINTNET2,PointCNN,DBLP:journals/corr/abs-1908-04512,LiuFXP19,PointTransformer,PAConv,WangSLSBS19} and voxel-based~\cite{sparseconv,Minkowski,yan2018second,CenterNet} networks. In both cases, the state-of-the-art models consist of an encoder network to  gradually downsample the points/voxels by sampling algorithms / strided convolution, a decoder network to propagate the features of the subsampled points/voxels into the original ones, and a task-specific head for making predictions. 
We call these methods \textbf{E}ncoder-\textbf{D}ecoder paradigm (ED-Paradigm) models.
Due to the downsampling-upsampling design, ED-Paradigm models extract features for some fixed positions appearing in the downsampling process.

In this paper, we propose a novel paradigm for 3D understanding tasks, namely \textbf{E}mbedding-\textbf{Q}uerying paradigm (EQ-Paradigm). Compared to the ED-Paradigm which extracts features for fixed positions, EQ-Paradigm enables feature generation for any position in the 3D scene. 
Thus, the EQ-paradigm is a generalization of the ED-Paradigm. 
Any ED-Paradigm model has an EQ-Paradigm counterpart. 
An EQ-Paradigm model consists of three stages: an \textbf{E}mbedding stage, a \textbf{Q}uerying stage and a task-specific head.
The embedding stage can be implemented with any feature extraction architecture, including both voxel- and point-based networks, regardless of tasks and heads. We use the embedding network to extract \textit{support features} for the following stages.
The querying stage then takes a set of artificially designated positions as \textit{query positions} and generates their intermediate representation, \ie, \textit{Q-representation}, based on the support features.
Note that the query positions could be any point in the contiguous 3D space, thus enabling feature generation for any location. 
We further present a novel querying stage network called \textit{Q-Net} to effectively extract Q-representation. Afterward, a task head is employed for generating predictions based on the Q-representation.

Due to the flexibility in query position designation, the EQ-Paradigm is a unified query-based paradigm that can easily combine any state-of-the-art 3D backbone networks with different task heads without extra efforts (Figure~\ref{fig:teaser}), which gives a lot of freedom in the head design.
For example, SSD head \cite{SSD} designed for voxel-based detectors \cite{yan2018second,VOXELNET} can be applied with a point-based embedding network under EQ-Paradigm;  an EQ-Paradigm segmentation model can directly obtain point-wise features based on a voxel-based embedding network \cite{Minkowski,sparseconv}; also, an EQ-Paradigm version of PVRCNN~\cite{PVRCNN} is able to directly generate proposal grid features from the voxel-based backbones for the following detection head. This greatly increases the flexibility of model design for different tasks. 

We evaluate our EQ-Paradigm on multiple 3D understanding tasks including object detection \cite{shi2018pointrcnn,yan2018second,PVRCNN,VoteNet,GroupFree}, semantic segmentation \cite{sparseconv, Minkowski} and shape classification~\cite{POINTNET2}.  
Our experiments show that our EQ-Paradigm and Q-Net can be well integrated with any state-of-the-art models regardless of tasks, backbone architectures and head designs, with a consistent performance improvement.
Our primary contributions are listed below.

\begin{itemize}\vspace{-0.05in}

\item We propose an \textbf{E}mbedding-\textbf{Q}uerying paradigm for 3D point cloud understanding. It is a unified query-based paradigm enabling the combination of arbitrary point- or voxel-based networks with different task heads. 

\vspace{-0.05in}
\item We present a novel querying stage network, Q-Net, to extract the intermediate Q-representation, \ie, query features, for the artificially designated query positions.

\vspace{-0.05in}
\item We integrate our EQ-Paradigm and Q-Net into multiple state-of-the-art 3D networks for different tasks and observe a consistent performance improvement from extensive experiments.
\end{itemize}

\section{Related Works}
\myparagraph{ED-Paradigm} 
ED-Paradigm models are widely applied in computer vision. They consist of an encoder network to extract high-level semantic features, a decoder network for feature propagation and a task head to perform predictions. U-Net \cite{UNET} is a classical ED-Paradigm network using a U-like architecture to deal with biomedical image segmentation. It inspires many modern works on 2D pixel-level tasks including semantic segmentation \cite{FCN,PSPNET,deeplabv3plus2018,DEEPLAB}, super resolution (SR) \cite{wang2019edvr} and matting \cite{DIM,SIM,forte2020fbamatting}. In 3D tasks, it is also a mainstream paradigm used in object detection \cite{FPOINTNET,VOXELNET,yan2018second,lang2018pointpillars} and semantic segmentation \cite{PointTransformer,LiuFXP19,WuQL19,DBLP:journals/corr/abs-1908-04512,DBLP:journals/corr/abs-1904-08889}. 

\myparagraph{Point-based 3D Architectures}
Point-based 3D models deal with raw point clouds, which extract sparse point features and downsample the point cloud in encoder networks, propagate features to original points by decoders, and make predictions by task-specific heads. PointNet++ \cite{POINTNET2} is the most popular and fundamental point-based backbone and has been widely applied in many point-based models \cite{yang2018ipod,yang3DSSD20,shi2018pointrcnn,FPOINTNET,wang2019frustum,Qi2019Votenet,GroupFree}.
These models utilize a series of set-abstraction layers as their encoders and multiple feature propagation layers as decoders. Some models focus on developing elegant heads to leverage the  sparse point features. For example, \cite{FPOINTNET} proposes F-PointNet for amodal 3D box estimation. \cite{shi2018pointrcnn} and \cite{yang2019std} develop canonical 3D bounding box refinement and PointsPool layer,  respectively. 
Other point-based backbones \cite{PointWeb,PointCNN,SpiderCNN} focus on improving the feature aggregation operation in PointNet++ by introducing graph convolutions \cite{SimonovskyK17, WangSS18}, convolution-like operations \cite{DBLP:journals/corr/abs-1904-08889,PAConv,WuQL19} or transformer structure \cite{PointTransformer,PointFormer}. With raw point cloud without data conversion, a point-based model can extract features with accurate relative positions and structural information, but is limited in dealing with large-scale point cloud due to the high time complexity of some operators like ball query and farthest point sampling.

\myparagraph{Voxel-based 3D Architectures}
\label{par:ptbased_voxelbased_comp}
Voxel-based methods first divide raw point cloud into regular voxels, then apply convolutional neural networks (CNNs) composed of sparse \cite{sparseconv,sparseconvold,Minkowski} or dense \cite{VOXELNET} convolutions as their encoder and decoder networks to extract voxel features.
Still, voxel-based models are widely applied in various methods \cite{VOXELNET, lang2018pointpillars,yan2018second,shi2019part,PVRCNN,Minkowski} on different tasks. Compared to point-based architectures, voxel-based methods reduce a large number of redundant points in the same voxels but sacrifice the data precision during the voxelization process. However, voxel-based methods are able to deal with the large-scale scenario.
In this paper, we propose an EQ-Paradigm for enabling models on different tasks to easily switch between these two backbone architectures and providing great flexibility in head design.

\section{EQ-Paradigm}
\vspace{-0.05in}
We first give an overview of the EQ-paradigm in this section, then elaborate our novel querying stage design, Q-Net, in the next section.

\subsection{Overview}
\begin{figure}[t]
  \centering
  \includegraphics[width=1.0\linewidth]{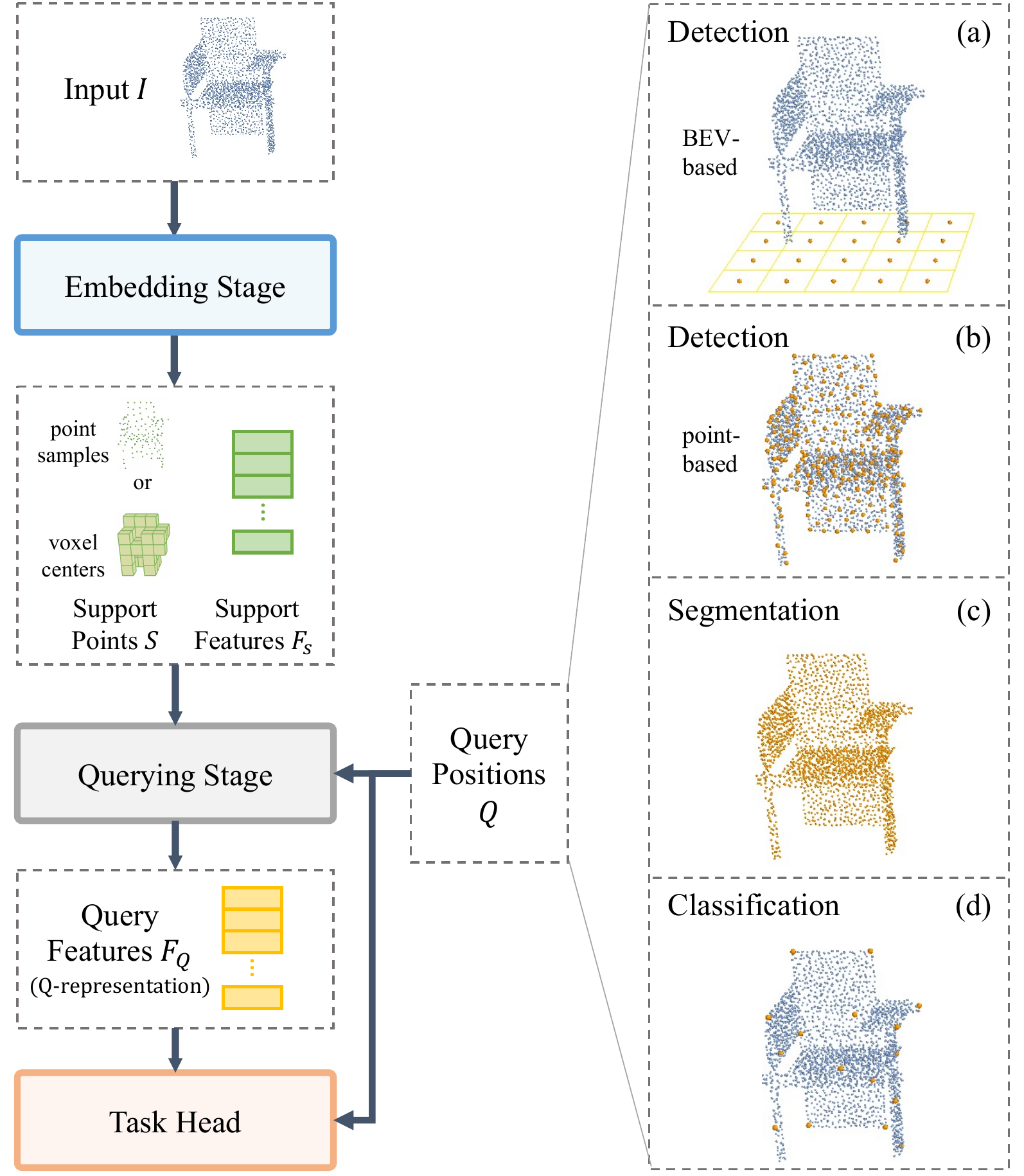}\\
  \vspace{-0.1in}
  \caption{
  An overview of our EQ-Paradigm. Given an input point cloud $I$, a set of support features $F_S$ for support points $S$ are generated in the embedding stage. The support points (marked in \textcolor[rgb]{0.49,0.67,0.33}{\textbf{green}}) can be voxel centers or point samples for voxel- or point-based embedding networks, respectively. The querying stage network generates the query features $F_Q$ (also known as Q-representation) used in the task head for query positions $Q$ based on $S$ and $F_S$. The query positions $Q$ (marked in \textcolor[rgb]{1.0,0.75,0}{\textbf{yellow}}) for different tasks and heads are shown in (a)-(d).
  }
  \label{fig:query_position}
  \vspace{-0.2in}
\end{figure}

As shown in Figure~\ref{fig:query_position}, our EQ-Paradigm has three stages: an \textbf{E}mbedding stage, a \textbf{Q}uerying stage and a task head. 
The embedding stage extracts features from the input point cloud $I\in\mathbb{R}^{N\times3}$. 
We take those features as support features $F_S$ for the following querying stage. The corresponding 3D positions of $F_S$ are denoted as support points $S\in\mathbb{R}^{n\times3}$. 
The querying stage is then responsible for generating Q-representation, that is, query features $F_Q$ for query positions $Q\in\mathbb{R}^{m\times3}$ based on support points $S$ and support features $F_S$. Notably, $Q$ is not required to be a subset of $I$. Instead, a query point is expected to be any position in the continuous 3D space, not just these input points collected by a 3D sensor. We provide a novel querying stage design called Q-Net in Section~\ref{sec:qnet}.
Finally, the task head produces predictions based on query positions $Q$ and features $F_Q$. Formally, our EQ-Paradigm can be formulated as:
\begin{equation} \label{eq:paradigm}
	\begin{aligned}
		F_S, S &= \text{Embedding}(I),\\
		F_Q &= \text{Querying}(F_S, S, Q), \\
		O &= \text{Head}(F_Q, Q),
	\end{aligned}
\end{equation}
where $O$ indicates the final outputs for specific tasks.

\subsection{Embedding Stage}
In the EQ-Paradigm, the feature extraction network in the embedding stage can be any 3D network, including voxel-based networks with voxelized inputs \cite{sparseconv,sparseconvold, Minkowski} and point-based networks with raw point clouds \cite{LiuFXP19,POINTNET,POINTNET2,WuQL19},  independent of tasks and heads. The goal of the embedding stage is to generate support points $S$ and support features $F_S$. For point-based embedding networks, the support points $S$ are usually a subsample of the input point cloud $I$, depending on the downsampling strategy (\eg, farthest point sampling) of the network;  in the voxel-based situation, the downsampling is usually made by strided convolutions, and we take the downsampled voxel centers as $S$. 

As mentioned in Section~\ref{par:ptbased_voxelbased_comp}, a voxel-based backbone is able to deal with large-scale point cloud scenarios, while a point-based backbone can extract more precise structural information. In the EQ-Paradigm, a model can arbitrarily specify its embedding stage network according to the practical demand, which brings flexibility in model design.

\subsection{Querying Stage}
\vspace{-0.05in}
The querying stage is utilized for extracting query features $F_Q$ for a set of manually designated query positions $Q$ from support features $F_S$ and their positions $S$. The queried features are then sent to the task-specific head for generating final predictions. 

The key aspect of the querying stage lies in the selection of query positions according to different tasks and head designs, as illustrated in Figure~\ref{fig:query_position} and the following. 
\vspace{-0.1in}
\begin{itemize}
    \item \textit{Query positions in detection.} To deploy an SSD \cite{SSD, yan2018second} head in an outdoor 3D object detection model, query positions are selected to be the pixel centers within the target Bird-Eye-View (BEV) map (Figure~\ref{fig:query_position}(a)).
    To utilize point-based heads proposed in \cite{shi2018pointrcnn,yang3DSSD20,VoteNet}, query positions are subsampled points from the raw input point cloud by uniform or farthest point sampling (Figure~\ref{fig:query_position}(b)).
    \vspace{-0.1in}
    \item \textit{Query positions in segmentation.} In semantic segmentation, query positions are the points requiring point-wise class predictions in a 3D scene (Figure~\ref{fig:query_position}(c)). Usually, the whole input point cloud $I$ is taken as $Q$.
    \vspace{-0.1in}
    \item \textit{Query positions in classification.} In classification, $Q$ can be the shape center to produce a representative feature for the classifier, or can also be multiple uniformly-distributed positions indicating an object's different parts to vote the category (Figure~\ref{fig:query_position}(d)). In this paper, we intend to vote an object's category by using 16 sampled points as query positions.
\end{itemize}
\vspace{-0.1in}
The querying stage is agnostic of the embedding network type and has great flexibility in query position selection. The point or voxel features extracted in the embedding stage can be well propagated to the query positions required by different tasks and heads.
Also, for a specific task head, it is possible to switch the point- or voxel-based embedding networks depending on which representation is better for the head. This is valuable in tasks like detection, where the head and backbone designs are both important, as shown in the ablation study in Section~\ref{sec:unify_representation}.

\begin{figure*}[t]
	\centering
	\includegraphics[width=1.0\linewidth]{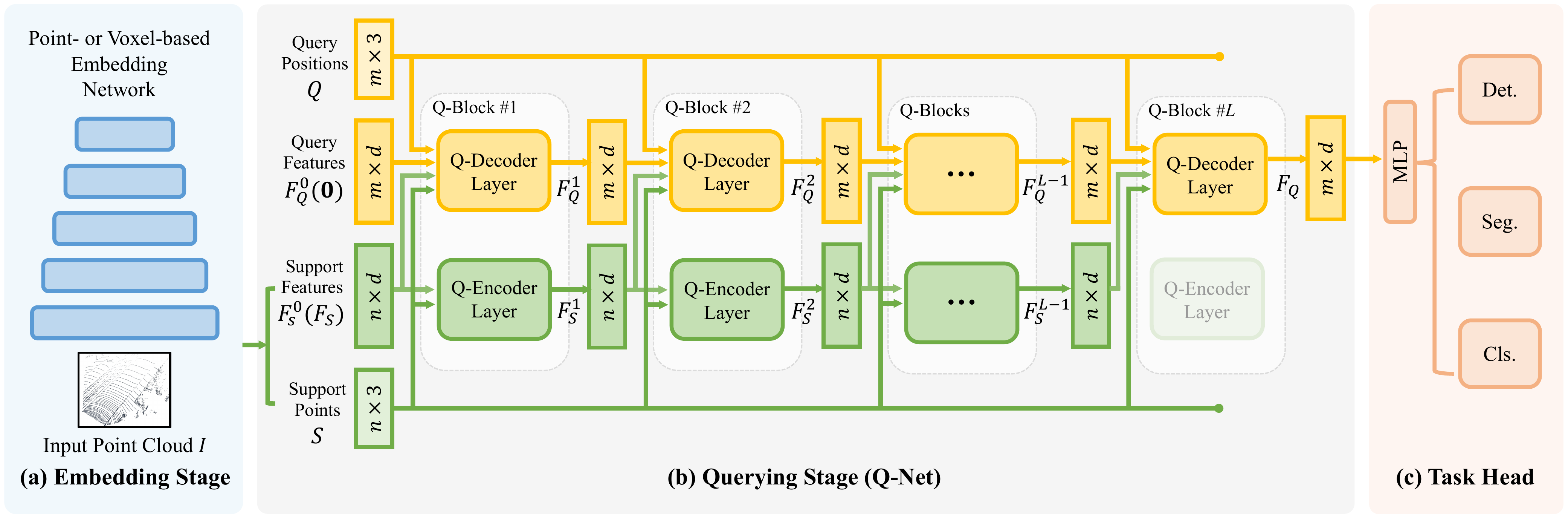}\\
	\vspace{-0.1in}
	\caption{Illustration of our Q-Net in EQ-Paradigm. 
	Taking as input the support points $S$ and support features $F_S$, Q-Net generates query features $F_Q$ for query positions $Q$. 
	Q-Net consists of $L$ consecutive Q-Blocks, each containing a Q-Encoder layer for updating support features and a Q-Decoder layer for refining query features.
	We initialize $F_Q^0$ by $\mathbf{0}$ and take $F_S$ as the initial support features $F_S^0$. 
	}
	\label{fig:qnet}
	\vspace{-0.2in}
\end{figure*}

\section{Q-Net}
\label{sec:qnet}
In this section, we propose a novel querying stage network, Q-Net, based on the transformer structure~\cite{AttentionIsAllYouNeed,VIT} to extract Q-representations, \ie, query features $F_Q$. 
Recently, transformer models have shown great potential in the field of 2D computer vision~\cite{VIT,DeIT,ConvIT,SwinTransformer,FocalTransformer,DETR} as well as 3D tasks~\cite{misra2021-3detr,PointTransformer,PointFormer,3DMAN}. 
Here, we develop our Q-Net based on the transformer to effectively generate features for query positions due to the flexible receptive field and strong representation ability of transformer layers. 
Note that the transformer mechanism is well-suited for the querying stage because the attention operator with positional encoding both contributes a global perspective and considers the relative positions between points, which fits our demands of feature generation for flexible query positions well.
Figure \ref{fig:qnet} shows the architecture of Q-Net.

\subsection{Q-Block}
In general, Q-Net is a stack of $L$ Q-Blocks. Each Q-Block has four input elements. For the $l$-th block, the four inputs are query positions $Q$, support points $S$, query features $F_Q^{l-1}$ and support features $F_S^{l-1}$, where $F_Q^{l-1}$ and $F_S^{l-1}$ are the outputs of the $(l-1)$-th block. For the first Q-Block, we initialize $F_Q^0$ by $\mathbf{0}$. Since query positions $Q$ are arbitrarily designated in a 3D scene and not required as a subset of input point cloud $I$, initializing their features with zeros does not introduce any inductive bias. Meanwhile, $F_S^0$ is initialized by the support features $F_S$ extracted in the embedding stage. Based on the initialization, these $L$ Q-Blocks iteratively update the query and support features. Here, $L$ is set to 3 in our implementation. Ablation study on $L$ can be found in the supplementary materials.

Each Q-Block utilizes two layers, a Q-Encoder layer and a Q-Decoder layer, to update the support features and refine the query features, respectively. The support features are updated to encode richer global semantic information, thus benefiting the query feature refinement. We abandon the Q-Encoder layer in the last Q-Block, since we do not need updated support features anymore without the next Q-Decoder layer. The output of the last Q-Block is the final query features $F_Q$, which are fed into the task head for making predictions. 
Formally, the Q-Block is depicted as:
\begin{equation} \label{eq:decoder_overall}
\begin{aligned}
F_Q^l &= \text{Q-Decoder}(Q, F_Q^{l-1}, S, F_S^{l-1}), \\
F_S^l &= \text{Q-Encoder}(S, F_S^{l-1}).
\end{aligned}
\end{equation}
We follow the original transformer~\cite{AttentionIsAllYouNeed} to build our Q-Encoder and Q-Decoder layers. We adopt the transformer encoder layer as our Q-Encoder layer, while the Q-Decoder layer is adapted from the transformer decoder layer.

\myparagraph{Q-Encoder Layer} We use the Q-Encoder layer to update the support features. The architecture of our Q-Encoder layer follows the widely-used transformer encoder layer which consists of two major components: an attention layer (Attention) and a feed-forward network (FFN). We formulate the Q-Encoder layer as: 
\begin{equation}
\begin{aligned}
	\hat{F}_S^l = &\ \text{Attention}(S, F_S^{l -1}, S, F_S^{l-1}) + F_S^{l -1}, \\
	F_S^l = &\ \text{FFN}(\hat{F}_S^l) + \hat{F}_S^l. \\
\end{aligned}
\end{equation}
The attention layer here is a classical $\mathbf{qkv}$-based multi-head self-attention~\cite{AttentionIsAllYouNeed}, where $\mathbf{q}$, $\mathbf{k}$ and $\mathbf{v}$ are all from the support features $F_S^{l -1}$. 
We use LayerNorm~\cite{layernorm} to normalize features before each Attention and FFN module.

\myparagraph{Q-Decoder Layer}
The Q-Decoder layer generates enhanced feature representations for query positions. 
Different from the transformer decoder layer, in the Q-Decoder layer, we do not apply self-attention on query features but directly adopt the cross-attention layer to generate query features from the support features, which is formulated as:
\begin{equation}
\begin{aligned}
	\hat{F}_Q^l = &\ \text{Attention}(Q, F_Q^{l -1}, S, F_S^{l-1}) + F_Q^{l -1}, \\
	F_Q^l = &\ \text{FFN}(\hat{F}_Q^l) + \hat{F}_Q^l, \\
\end{aligned}
\end{equation}
where the attention layer is a $\mathbf{qkv}$-based multi-head cross-attention, in which $\mathbf{q}$ is from the query features while $\mathbf{k}$ and $\mathbf{v}$ are from the support features. 
Removing the self-attention in the conventional transformer decoder layer keeps the independence of query positions, that is, the query feature of a query position only depends on its relationship with the support points/features but not with other query positions/features, thus providing more freedom in the choice of query positions. For example, we can query the features of only parts of interest in the whole scene. The ablation study in Section~\ref{sec:ab_qdecoder} shows the advantages of this design.

\myparagraph{Attention Layer} The Attention layer, formulated as: 
\vspace{-1mm}
\begin{equation} \label{eq:transformer_overall}
	\begin{aligned}
		\tilde{F}_Y = \text{Attention}(Y, F_Y, X, F_X),
	\end{aligned}
\end{equation}
plays a fundamental role in a Q-Block. It leverages $m$ target positions $Y\in\mathbb{R}^{m\times3}$ with features $F_Y\in\mathbb{R}^{m\times d}$ and $n$ source positions $X\in\mathbb{R}^{n\times3}$ with features $F_X\in\mathbb{R}^{n\times d}$ to obtain new target features $\tilde{F}_Y\in\mathbb{R}^{m\times d}$. Here, $d$ denotes the channel number of features.
A $\mathbf{qkv}$-based attention layer can be viewed as applying attention weights on the source features $F_X$ for computing new target features. 
Here, we describe the single-head calculation for clarity.
The computation of the $i$-th new target feature $\tilde{F}_Y^{(i)}$ is formulated as:
\begin{equation} \label{eq:attention_whole}
\begin{aligned}
\tilde{F}_Y^{(i)} = \mathcal{A}^{(i)}(F_XW_\mathbf{v} + B_\mathbf{v}^{(i)}).
\end{aligned}
\end{equation}
The attention weight $\mathcal{A} \in \mathbb{R}^{m \times n}$ is obtained by utilizing a softmax function on the result of dot product between target features $F_Y$ and source features $F_X$:
\begin{equation} \label{eq:attention_weight}
\begin{aligned}
\mathcal{A} = \text{SoftMax}\left(\frac{(F_YW_\mathbf{q})(F_XW_\mathbf{k})^T + B_{\mathbf{qk}}}{\sqrt{d}}\right).
\end{aligned}
\end{equation}
$W_\mathbf{q}$, $W_\mathbf{k}$ and $W_\mathbf{v}$ are weights of the linear layers for $\mathbf{q}$, $\mathbf{k}$ and $\mathbf{v}$, respectively.
Also, in our Q-Block, we apply two types of relative positional encoding. The first one $B_{\mathbf{v}}\in\mathbb{R}^{m\times n \times d}$ in Eq. \eqref{eq:attention_whole} is for providing relative geometric information in the value vectors. The second one $B_{\mathbf{qk}} \in \mathbb{R}^{m \times n}$ in Eq. \eqref{eq:attention_weight} encodes the Euclidean positional difference between the target $Y$ and source $X$ in the attention weights.

\myparagraph{Relative Positional Encoding}
Relative positional encoding is an indispensable component in our Q-Net. Unlike previous transformer structures~\cite{DETR, AttentionIsAllYouNeed} that adopt input features with effective semantic and position information, we initialize query features $F_Q^0$ by $\mathbf{0}$ in the first Q-Block, which avoids introducing inductive biases but provides no effective information. Hence, at the beginning of our Q-Net, query positions are the only hints for generating query features from support points and support features.
Meanwhile, it is not optimal to update query features in the first block only depending on the coordinate difference between query and support points, since it makes no difference in attention weights for object points with the same relative position but in various scales and shapes. Inspired by \cite{iRPE,RPE}, we adopt contextual relative positional encoding which fits our Q-Block well.

Compared to bias-mode relative positional encoding \cite{SwinTransformer,iRPE,FocalTransformer}, contextual relative positional encoding considers the interactions of positional embeddings with the $\mathbf{q}$, $\mathbf{k}$, $\mathbf{v}$ features, making the relative positional encoding automatically adapt to features with different contextual information. Hence, it can produce various responses for points in objects with diverse scales and shapes, even when some point pairs share the same relative positional difference. 
We provide the details of our relative positional encoding strategy and its effect in the supplementary materials. 

\myparagraph{Local Attention}
When the numbers of target $m$ and source $n$ are large, \ie, 40k, applying global attention on them is extremely GPU memory-consuming, since the attention weight $\mathcal{A}\in\mathbb{R}^{m\times n}$ is too large to store. 
To address this issue, we instead apply local attention in our Q-Net inspired by \cite{PointTransformer,PointFormer}. Specifically, for each target point, we figure out its $K$ nearest neighbors (KNN) in source points according to Euclidean distances and then compute attention only on these neighbors. In this way, the size of attention weight $\mathcal{A}$ is greatly reduced to $m \times K$ as $K$ is far smaller than $n$.

\begin{figure}[t]
  \centering
  \includegraphics[width=1.0\linewidth]{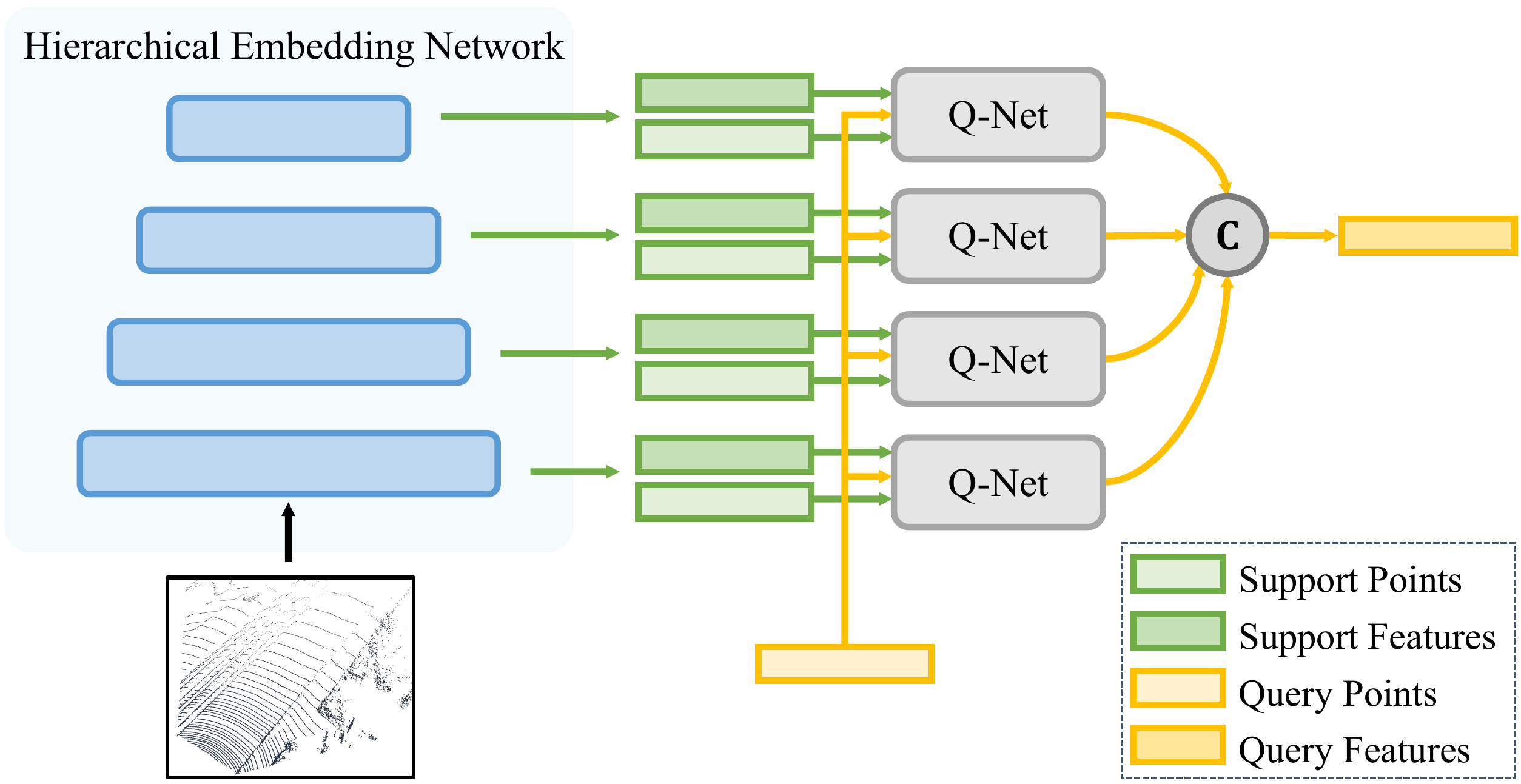}\\
  \vspace{-0.1in}
  \caption{The hierarchical extension of our Q-Net.}
  \label{fig:hierarchical}
  \vspace{-0.2in}
\end{figure}

\vspace{-0.05in}
\subsection{Hierarchical Q-Net}
\vspace{-0.05in}
Hierarchical multi-level architecture is proven to be essential for 3D tasks~\cite{POINTNET2, Minkowski} considering the diversity in 3D scene scales and object sizes. Especially for a point-wise prediction task like semantic segmentation, the multi-level features are of great importance in producing state-of-the-art results~\cite{Minkowski, PointTransformer}, since the fine-grained features are needed to make detailed per-point segmentation. 
Therefore, we develop a hierarchical Q-Net for exploiting multi-level features. As illustrated in Figure \ref{fig:hierarchical}, we apply a series of Q-Nets on support features from multiple levels of the hierarchical embedding network and concatenate the query features from different levels to generate final predictions.

\begin{table}[t]
	\centering \addtolength{\tabcolsep}{-1pt}
	\footnotesize
	\begin{tabular}{l|cc|cc}
		\hline
		\multicolumn{1}{l|}{\multirow{2}{*}{Method}} & 
		\multicolumn{2}{c|}{\multirow{1}{*}{ScanNet}} & 
		\multicolumn{2}{c}{\multirow{1}{*}{S3DIS}} \\
		\cline{2-5}
		\multicolumn{1}{l|}{\multirow{0}{*}{}} & 
		\multicolumn{1}{c}{\multirow{1}{*}{Validation}} & 
		\multicolumn{1}{c|}{\multirow{1}{*}{Test}} & 
		\multicolumn{1}{c}{\multirow{1}{*}{Area 5}} & 
		\multicolumn{1}{c}{\multirow{1}{*}{6-fold}}\\
		\hline
		\hline
		PointNet \cite{POINTNET} & - & - & 41.1 & 47.6 \\
		\hline
		PointNet++ \cite{POINTNET2} & - & 33.9 & - & - \\
		\hline
		PointCNN~\cite{PointCNN} & - & 45.8 & 57.3 & 65.4 \\
		\hline
		PointWeb~\cite{PointWeb} & - & - & 60.3 & 66.7 \\
		\hline
		PointEdge~\cite{pointedge} & 63.4 & 61.8 & 61.9 & 67.8 \\
		\hline
		PointConv~\cite{WuQL19} & 61.0 & 66.6 & - & - \\
		\hline
		PointASNL~\cite{pointasnl} & 66.4 & 66.6 & 62.6 & 68.7 \\
		\hline
		KPConv~\cite{DBLP:journals/corr/abs-1904-08889} & 69.2 & 68.6 & 67.1 & 70.6 \\
		\hline
		FusionNet~\cite{fusionnet} & - & 68.8 & 67.2 & - \\
		\hline
		SparseConvNet~\cite{sparseconv} & - & 72.5 & - & - \\
		\hline
		MinkowskiNet~\cite{Minkowski} & 72.2 & 73.6 & 65.4 & - \\
		\hline
	    PAConv~\cite{PAConv} & - & - & 66.6 & 69.3 \\
		\hline
		PointTransformer \cite{PointTransformer} & - & - & 70.4 & 73.5\\
		\hline
		\hline
		Sparse U-Net (Baseline) & 72.9 & - & 66.9 & 72.6\\
		\hline
		Sparse EQ-Net (Ours) & \textbf{75.3} & \textbf{74.3} & \textbf{71.3} & \textbf{77.5}\\
		\hline
		\rowcolor{LightCyan}
		\textit{Improvement} & \textit{+2.4} & - & \textit{+4.4} & \textit{+4.9}\\
		\hline
	\end{tabular}
	\vspace{-0.1in}
	\caption{Semantic segmentation results on mIoU(\%) of our method and other 3D networks on ScanNet and S3DIS. The Sparse U-Net is our re-implemented version of SparseConvNet.}
	\label{tab:semseg}
	\vspace{-0.1in}
\end{table}

\begin{table}[t]
   \centering \addtolength{\tabcolsep}{-1pt}
   \footnotesize
   \begin{tabular}{l|l|c|c}
       \hline
       \multicolumn{1}{l|}{\multirow{2}{*}{Method}} & 
       \multicolumn{1}{l|}{\multirow{2}{*}{Network}} & 
       \multicolumn{1}{c|}{\multirow{1}{*}{mAP}} & 
       \multicolumn{1}{c}{\multirow{1}{*}{mAP}} \\
       \multicolumn{1}{l|}{\multirow{0}{*}{}} & 
       \multicolumn{1}{l|}{\multirow{0}{*}{}} & 
       \multicolumn{1}{c|}{\multirow{1}{*}{@0.25}} & 
       \multicolumn{1}{c}{\multirow{1}{*}{@0.5}} \\
       \hline
       \hline
       \multicolumn{1}{l}{\multirow{1}{*}{\bf ScanNetV2}} & 
       \multicolumn{1}{l}{\multirow{1}{*}{}} & 
       \multicolumn{1}{c}{\multirow{1}{*}{}} & 
       \multicolumn{1}{c}{\multirow{1}{*}{}} \\
       \hline
       \hline
       VoteNet \cite{VoteNet} & PointNet++ & 58.6 & 33.5 \\
       VoteNet$^+$ & PointNet++ & 62.9 & 39.9 \\
       VoteNet (Ours) & EQ-PointNet++ & \textbf{64.3} & \textbf{45.4} \\
       \hline
       GroupFree \cite{GroupFree} & PointNet++ (L6, O256) & 67.3 & 48.9 \\
       GroupFree$^+$ & PointNet++ (L6, O256) & 66.3 & 47.8 \\
       GroupFree (Ours) & EQ-PointNet++ (L6, O256) & \textbf{68.0} & \textbf{50.0} \\
       \hline
       \hline
       \multicolumn{1}{l}{\multirow{1}{*}{\bf SUN RGB-D}} & 
       \multicolumn{1}{l}{\multirow{1}{*}{}} & 
       \multicolumn{1}{c}{\multirow{1}{*}{}} & 
       \multicolumn{1}{c}{\multirow{1}{*}{}} \\
       \hline
       \hline
       VoteNet \cite{VoteNet} & PointNet++ & 57.7 & 32.9 \\
       VoteNet$^+$ & PointNet++ & 59.1 & 35.8 \\
       VoteNet (Ours) & EQ-PointNet++ & \bf 60.5 & \bf 38.5 \\
       \hline
       \hline
   \end{tabular}
   \vspace{-0.1in}
   \caption{Performance of different methods with PointNet++ and EQ-PointNet++ on ScanNetV2 and SUN RGB-D datasets. $^+$ denotes the models reproduced by MMDetection3D \cite{mmdet3d2020}.}
   \label{tab:indoordet_main}
   \vspace{-0.2in}
\end{table}

\begin{table*}[t]
   \centering 
   \setlength{\tabcolsep}{3mm}{
   \footnotesize
   \begin{tabular}{c||l||c|c|c||c|c|c||c|c|c}
       \hline
       \multicolumn{1}{c||}{ \multirow{2}{*}{Set}} &
       \multicolumn{1}{c||}{ \multirow{2}{*}{Method}} &
       \multicolumn{3}{|c||}{Car (\%)} & \multicolumn{3}{|c||}{Pedestrian (\%)} & \multicolumn{3}{|c}{Cyclist (\%)} \\ \cline{3-11}
       \multicolumn{1}{c||}{} & \multicolumn{1}{c||}{} & \multicolumn{1}{|c|}{Easy} & \multicolumn{1}{|c|}{Moderate} & \multicolumn{1}{|c||}{Hard} & \multicolumn{1}{|c|}{Easy} & \multicolumn{1}{|c|}{Moderate} & \multicolumn{1}{|c||}{Hard} & \multicolumn{1}{|c|}{Easy} & \multicolumn{1}{|c|}{Moderate} & \multicolumn{1}{|c}{Hard} \\
       \hline
       \hline
      \multicolumn{1}{c||}{ \multirow{7}{*}{Val}} & SECOND \cite{yan2018second} & 90.85  & \bf 81.66  & 78.57 & 56.07 & 51.12 & 46.14 & 83.06 & 66.69 & 63.02 \\
      {} & EQ-SECOND (Ours) & \bf91.74 & 81.49 & \bf 78.62 & \bf57.48 & \bf53.64 & \bf49.55 & \bf85.01 & \bf67.13 & \bf63.34 \\
      \cline{2-11}
      {} & PointRCNN \cite{shi2018pointrcnn} & 91.35 & 80.25 & 77.84 & 61.19 & 54.33 & 47.43 & 89.77 & \bf 71.55 & \bf 67.20 \\
      {} & EQ-PointRCNN (Ours) & \bf91.80 & \bf84.00 & \bf82.29 & \bf64.80 & \bf58.36 & \bf52.55 & \bf91.23 & 71.09 & 66.35 \\
      \cline{2-11}
      {} & PVRCNN \cite{PVRCNN} & 92.07 & 84.75 & 82.46 & 62.32 & 54.42 & 49.81 & 90.39 & 70.42 & 65.99 \\
      {} & EQ-PVRCNN$^\dag$ (Ours) & \bf 92.63 & 85.41 & 82.97 & 66.78 & 59.23 & 54.34 & \bf 93.34 & \bf 75.71 & \bf 71.11 \\
      {} & EQ-PVRCNN$^\S$ (Ours) & 92.52 & \bf 85.61 & \bf 83.13 & \bf 69.95 & \bf 62.55 & \bf 56.51 & 91.51 & 74.02 & 69.46 \\
      \hline
      \hline
      \multicolumn{1}{c||}{ \multirow{2}{*}{Test}} & PVRCNN \cite{PVRCNN} & \bf 90.25 &	81.43 &	76.82 & 52.17 & 43.29 &	40.29 & 78.60 &	63.71 &	57.65 \\
      {} & EQ-PVRCNN$^\S$ (Ours) & 90.13 & \bf 82.01 & \bf 77.53 & \bf 55.84 & \bf 47.02 &	\bf 42.94 & \bf 85.41 & \bf 69.10 & \bf 62.30 \\ 
      \hline
   \end{tabular}}
   \vspace{-0.1in}
   \caption{Performance comparison on the KITTI val and test sets.
   }\label{tab:kitti_main}
   \vspace{-0.2in}
\end{table*}

\section{Experiments}
\vspace{-0.05in}
We conduct experiments on four popular 3D tasks: semantic segmentation, indoor object detection, outdoor object detection and shape classification. Implementation details of training schedule, hyper-parameters and network structure can be found in the supplementary materials.

\subsection{Semantic Segmentation}
\vspace{-0.1in}
\myparagraph{Datasets } For the task of point cloud semantic segmentation, we use two competitive and popular datasets, ScanNetV2~\cite{SCANNET} and S3DIS~\cite{s3dis}, in our experiments. ScanNetV2 comprises 1,613 indoor scans (1,201/312/100 for train/val/test) with point-wise semantic labels in 20 object categories. S3DIS is composed of 271 point cloud scenes collected from six large-scale indoor areas, which are annotated with 13 semantic classes. For evaluation, we follow the commonly-used S3DIS dataset split~\cite{PointCNN, PointWeb, Minkowski} to test on Area 5 and train on other five areas, and also apply the 6-fold cross validation which takes each area as test set once. For the evaluation metrics, we adopt the mean Intersection-over-Union (mIoU). 

\myparagraph{Models } We utilize the voxel-based residual U-Net structure with sparse convolutions~\cite{Minkowski, sparseconv} as the baseline model in our experiments. The sparse U-Net follows the ED-Paradigm with an encoder network followed by a decoder one, which is demonstrated to be a powerful backbone structure in 3D segmentation. We develop our network with EQ-Paradigm based on the sparse U-Net by keeping the encoder as our embedding network and adopting the Q-Net to extract the features for each point. In the embedding stage, the input 3D volume is downsampled six times, thus providing multi-level support features. We use the center coordinates of the voxels as the support positions and apply the hierarchical Q-Net to fuse the multi-level features to get better feature representations for the query positions. The queried features are then fed into a classifier to produce point-wise semantic predictions.

\myparagraph{Experimental Results } We compare our EQ-Net with our baseline model, \ie, the sparse U-Net, and other 3D networks. The results are shown in Table~\ref{tab:semseg}. On both datasets, our method attains higher mIoU than the strong baseline models, with significant gains of 2.4\%, 4.4\% and 4.9\% on ScanNet validation set, S3DIS Area 5 and 6-fold, respectively. Also, compared with recent state-of-the-art 3D segmentation networks, our EQ-Net still achieves higher performance on these two datasets, showing the effectiveness of the EQ-Paradigm and our well-designed Q-Net in point-wise prediction tasks.

\subsection{Indoor Object Detection}
\vspace{-0.1in}
\myparagraph{Datasets } 
We evaluate our method on two popular datasets: ScanNetV2 \cite{SCANNET} and SUN RGB-D \cite{sunrgbd}. ScanNetV2 contains 1,513 scenes with bounding boxes labeled in 18 categories; SUN RGB-D includes 5k training scenes with bounding boxes in 10 classes. Evaluation metric is the mean Average Precision (mAP) with intersection-over-union (IoU) $0.25$ (mAP@0.25) and $0.5$ (mAP@0.5) following \cite{VoteNet}.

\myparagraph{Baseline Models}
We test our approach on VoteNet \cite{VoteNet} and GroupFree \cite{GroupFree} for ScanNetV2 dataset, and on VoteNet for SUN RGB-D dataset. All baseline models are publicly available at MMDetection3D \cite{mmdet3d2020} codebase. VoteNet is the classical indoor detector serving as the baseline model for all modern methods; GroupFree is the current state-of-the-art indoor detector.

\myparagraph{EQ-PointNet++}
PointNet++ \cite{POINTNET2} is the cornerstone of indoor 3D object detection. Modern methods \cite{VoteNet,GroupFree} all utilize it to extract sparse point features for their detection heads.
EQ-PointNet++ is the EQ-Paradigm version of PointNet++. It treats a stack of set-abstraction layers as its embedding stage similar to PointNet++ and applies a hierarchical Q-Net in its querying stage to extract query features with multi-level information. Query positions are 1,024 points obtained by applying furthest point sampling (FPS) on the raw input point cloud following \cite{VoteNet,GroupFree}. For all models, we replace their PointNet++ backbone networks by our EQ-PointNet++ networks.

\myparagraph{Experimental Results}
As shown in Table \ref{tab:indoordet_main}, models with EQ-PointNet++ achieve better performance on both datasets. Specifically, VoteNet with EQ-PointNet++ gains $5.5 \%$ and $2.7 \%$ mAP@0.5 improvements on ScanNetV2 and SUN RGB-D datasets respectively. On the state-of-the-art indoor detector GroupFree \cite{GroupFree}, our approach brings consistent performance improvements on mAP@0.25 and mAP@0.5 by $0.7 \%$ and $1.1 \%$ compared to the official results \cite{GroupFree} and by $1.7 \%$ and $2.2 \%$ compared to our reproduced results \cite{mmdet3d2020}. These experiments demonstrate our EQ-Paradigm and Q-Net can be well adapted into indoor detectors and boost their performance.

\subsection{Outdoor Object Detection}
\vspace{-0.1in}
\myparagraph{Datasets}
For outdoor detection, we conduct experiments on the widely adopted KITTI dataset \cite{KITTIDATASET1}. There are 7,481 training point clouds and 7,518 testing point clouds with three categories of ``Car'', ``Pedestrian'' and ``Cyclist''. Following \cite{VOXELNET}, we split the original KITTI training dataset to 3,717 images/scenes train set and 3,769 images/scenes val set. All ``AP'' results are calculated with 40 recall positions following the official KITTI protocol. 

\myparagraph{Baseline Models}
We select three outdoor detectors to demonstrate the superiority of our approach: SECOND \cite{yan2018second}, PointRCNN \cite{shi2018pointrcnn} and PVRCNN \cite{PVRCNN}. These methods with different heads require different query position designation. In SECOND, query positions are the pixel centers within the target BEV map; in PointRCNN, all points within the input point cloud serve as query positions; in PVRCNN, they can be coordinates either of keypoints (EQ-PVRCNN$^\S$) following the original PVRCNN design or of proposal grids straightforwardly (EQ-PVRCNN$^\dag$).

\myparagraph{Experimental Results}
As listed in Table \ref{tab:kitti_main}, our approach brings a consistent improvement on different detectors. Especially on PointRCNN, EQ-PointRCNN obtains significant improvements, \eg, $3.75 \%$ ``AP'' improvement on ``Car'' instances labeled as ``Moderate'' difficulty. Compared to the state-of-the-art model, PVRCNN, our approach achieves remarkable improvements on both KITTI val and test sets. Specifically, on the test set, EQ-PVRCNN$^\S$ attains $3.73 \%$ and $5.39 \%$ improvements on ``Pedestrian'' and ``Cyclist'' instances labeled as ``Moderate'' difficulty level.
These practical improvements indicate that EQ-Paradigm and Q-Net can be widely applied on any 3D outdoor detectors and deliver a sustained performance improvement. Meanwhile, by altering query positions, our approach can inspire some new designs on existing methods. As shown in Table \ref{tab:kitti_main}, by directly obtaining proposal grid features for box prediction to get rid of modules including voxel-set abstraction, predicted keypoint weighting, and RoI-grid Pooling in PVRCNN, EQ-PVRCNN$^\dag$ still achieves impressive performance improvement yet with concise head design.

\begin{table}[t]
   \centering \addtolength{\tabcolsep}{-1pt}
   \footnotesize
   \begin{tabular}{l|c | c }
       \hline
       Method & Input  & Accuracy (\%) \\
       \hline
       \hline
       PCNN \cite{PCNN} & 1k points & 92.3 \\
       RS-CNN (SSG) \cite{LiuFXP19} & 1k points & 92.4 \\
       PointCNN \cite{PointCNN} & 1k points & 92.5 \\
       KPConv \cite{DBLP:journals/corr/abs-1904-08889} & 1k points & 92.9 \\
       DGCNN \cite{WangSLSBS19} & 1k points & 92.9 \\
       InterpCNN \cite{DBLP:journals/corr/abs-1908-04512} & 1k points & 93.0 \\
       DensePoint \cite{DensePoint} & 1k points & 93.2 \\
       Grid-GCN \cite{GridGCN} & 1k points & 93.1 \\
       PosPool \cite{PosPool} & 5k points & 93.2 \\
       \hline
       SpecGCN \cite{SpecGCN} & 2k points+normal & 92.1 \\
       PointWeb \cite{PointWeb} & 1k points + normal & 92.3 \\
       SpiderCNN \cite{SpiderCNN} & 1k points+normal & 92.4 \\
       PointConv \cite{WuQL19} & 1k points+normal & 92.5 \\
       \hline
       \hline
       PointNet++ (SSG) & 1k points & 92.1 \\
       EQ-PointNet++ (SSG) & 1k points & \textbf{93.2} \\
      \hline
   \end{tabular}
   \vspace{-0.1in}
   \caption{Accuracy comparison on the ModelNet40 dataset.}
   \label{tab:modelnet40_main}
   \vspace{-0.1in}
\end{table}

\subsection{Shape Classification}
\myparagraph{Datasets}
We conduct classification experiments on ModelNet40 dataset \cite{modelnet40} which includes 9,843 training and 2, 468 testing meshed models in 40 categories.

\myparagraph{Models}
We employ EQ-PointNet++ as our classification model. Query positions are 16 points obtained by furthest point sampling on the input point cloud. In the recognition head, we deploy another set-abstraction layer to summarize the 16 query features for obtaining category prediction.

\myparagraph{Experimental Results}
As shown in Table \ref{tab:modelnet40_main}, EQ-PointNet++ surpasses the classification accuracy of PointNet++ with single-scale grouping (SSG) by $1.1 \%$. Compared with other classifiers \cite{WangSLSBS19}, EQ-PointNet++ still shows better performance, which demonstrates the generalization ability of EQ-Paradigm and the effectiveness of Q-Net.

\begin{table}[t]
   \centering \addtolength{\tabcolsep}{-1pt}
   \footnotesize
   \begin{tabular}{p{2.4cm} || c | c || c }
       \hline
       \multicolumn{1}{c||}{ \multirow{2}{*}{\makecell{Head}}} & 
       \multicolumn{2}{c||}{ \multirow{1}{*}{Embedding Network}}  & 
       \multicolumn{1}{c}{ \multirow{2}{*}{AP (\%)}}  \\ \cline{2-3}
       {} & \multicolumn{1}{c|}{ \multirow{1}{*}{voxel-based}} & \multicolumn{1}{c||}{ \multirow{1}{*}{point-based}} & {} \\
       \hline
       \hline
       \multicolumn{1}{c||}{ \multirow{3}{*}{\makecell{SECOND head \\ (voxel-based)}}} & $\surd$ & - & 81.49 \\
        & - & $\surd$ & 82.70 \\
        & $\surd$ & $\surd$ & 82.94 \\ 
       \hline
       \hline
       \multicolumn{1}{c||}{ \multirow{3}{*}{\makecell{PointRCNN head \\ (point-based)}}} & $\surd$ & - & 82.65 \\
        & - & $\surd$ & 84.00 \\
        & $\surd$ & $\surd$ & 84.38 \\
      \hline
   \end{tabular}
   \vspace{-0.1in}
   \caption{AP comparison on different head designs with point- and voxel-based embedding networks.}
   \label{tab:representation_comp}
   \vspace{-0.1in}
\end{table}

\subsection{Ablation Study}
\vspace{-2mm}
\myparagraph{Analysis on the EQ-Paradigm}
\label{sec:unify_representation}
In Table \ref{tab:representation_comp}, we verify the capacity of EQ-Paradigm in combining point- or voxel-based backbone networks with different task heads by adopting different embedding structures in voxel- and point-based detectors, SECOND \cite{yan2018second} and PointRCNN \cite{shi2018pointrcnn}.
Experiments are conducted on KITTI validation set with ``AP'' calculated on “Moderate” difficulty level in class ``Car''. We use the SparseConvNet in SECOND \cite{yan2018second} as the voxel-based embedding network, and PointNet++ without decoder in PointRCNN \cite{shi2018pointrcnn} as the point-based embedding network.

As illustrated in Table \ref{tab:representation_comp}, heads in SECOND and PointRCNN are both applicable on point- and voxel-based embedding stage networks and produce promising performance. This manifests the EQ-Paradigm unifies different 3D architectures.
Notably, SECOND with point-based embeddings achieves $1.21 \%$ improvement over its voxel-based baseline. This demonstrates that different architectures have unique advantages. For example, point-based architectures are able to extract more precise structural information.
Meanwhile, in Table \ref{tab:representation_comp}, we show that voxel- and point-based embedding networks can be simultaneously utilized in an EQ-Paradigm model to bring further improvement. These experiments demonstrate that EQ-Paradigm gives great flexibility in backbone and head selection and is able to combine strengths of points and voxels. 

\myparagraph{Analysis on the Hierarchical Q-Net} 
Multi-level features play an important role in recognition tasks~\cite{PSPNET,DEEPLAB,FPN}. Our EQ-Paradigm is naturally compatible with the multi-level scheme by simply employing multi-level features as support features in the querying stage. 
We accordingly design a simple yet effective hierarchical Q-Net structure. We validate the promotion from fusing multi-level information by conducting experiments on point cloud semantic segmentation, which calls for fine-grained features
to better segment the points on object boundaries. Table~\ref{tab:hierarchical_qnet} shows the effects of incorporating different levels of features in the querying stage on ScanNet validation set.
We start from the coarsest layer and gradually include more finer features in the querying process.
Continuous performance improvement is observed as the number of feature levels increases, manifesting the effectiveness of our hierarchical Q-Net.

\begin{table}[t]
	\centering \addtolength{\tabcolsep}{-1pt}
	\footnotesize
	\begin{tabular}{ c | c c c c c c}
		\hline
		No. of Levels & 1 & 2 & 3 & 4 & 5 & 6\\
		\hline
		mIoU (\%) & 58.4 & 64.1 & 68.3 & 71.9 & 74.2 & 75.3 \\
		\hline
	\end{tabular}
	\vspace{-0.1in}
	\caption{Effects of the number of feature levels in our hierarchical Q-Net. The experiments are conducted on ScanNet validation set.} 
	\label{tab:hierarchical_qnet}
	\vspace{-0.1in}
\end{table}

\begin{table}[t]
   \centering \addtolength{\tabcolsep}{-1pt}
   \footnotesize
   \begin{tabular}{p{1.3cm} |c || c | c || c }
       \hline
       \multicolumn{1}{c|}{ \multirow{2}{*}{Method}} & 
       \multicolumn{1}{c||}{ \multirow{2}{*}{SA}} &
       \multicolumn{2}{c||}{ \multirow{1}{*}{Query Position Selection}}  & 
       \multicolumn{1}{c}{ \multirow{2}{*}{AP (\%)}}  \\ \cline{3-4}
       {} & {} & \multicolumn{1}{c|}{ \multirow{1}{*}{train}} & \multicolumn{1}{c||}{ \multirow{1}{*}{test}} & {} \\
       \hline
       \hline
       \multicolumn{1}{c|}{ \multirow{4}{*}{EQ-SECOND}} & $\surd$ & patch & patch & 81.61 \\
       {} & $\surd$ & patch & random & 74.96 \\ \cline{2-5}
       {} & - & patch & patch & 81.49 \\
       {} & - & patch & random & 81.49 \\
      \hline
   \end{tabular}
   \vspace{-0.1in}
   \caption{AP comparison on EQ-SECOND utilizing Q-Decoder layer with or without self-attention (``SA'') layers.}
   \label{tab:qdecoder}
   \vspace{-0.1in}
\end{table}

\myparagraph{Analysis on the Q-Decoder}
\label{sec:ab_qdecoder}
In Q-Decoder layer, to make query positions independent of each other to allow arbitrary query position selection, we remove the self-attention layer for query points in the conventional transformer decoder layer. In Table \ref{tab:qdecoder}, we compare the performance of EQ-SECOND with and without the self-attention layer on different test modes. Both models are trained in ``patch'' mode, and tested in two modes, ``patch'' and ``random''. In ``patch'' mode, we split the target BEV map into patches with equal sizes, randomly select one patch at each iteration, and treat all pixel centers within the patch as query positions.
In ``random'' mode, we arbitrarily choose some pixel centers within the BEV map as query positions.

The self-attention layer encodes the relations among query positions, thus restricting the choice of query positions at test time. 
From the results in Table~\ref{tab:qdecoder}, we observe an AP drop of $6.65 \%$ on the model with self-attention layer when tested with randomly-selected query positions, demonstrating the great negative effect of self-attention layer to arbitrary query position selection.
On the contrary, our model free of self-attention enables arbitrary selection with no performance drop.
Another concern for the self-attention layer is that it brings very limited AP improvement ($0.12 \%$) but incurs a large computation overhead when dealing with a large number of query positions.

\vspace{-0.1in}
\section{Conclusion}
\vspace{-0.05in}
In this paper, we present a novel unified pipeline, the EQ-Paradigm, for 3D understanding tasks including object detection, semantic segmentation and classification. The EQ-Paradigm enables a free combination of 3D backbone architectures, heads and tasks. We achieve this by proposing a querying stage to transfer the support features extracted in the embedding stage to the positions required by heads and tasks. We further develop a dedicated Q-Net as the querying stage network, which is applicable for multiple state-of-the-art models on different tasks and brings performance improvement. In the future, we plan to generalize our EQ-Paradigm to other 3D tasks like scene completion and instance segmentation. We believe that the EQ-Paradigm could be a unified pipeline for various 3D tasks.

\section{Limitation}
Through EQ-Paradigm, we can easily combine different backbones and heads, which provides great flexibility in 3D model design. However, the choice of embedding stage network is still highly-dependent on the application scenarios. For example, due to the diverse scopes and point distributions on indoor and outdoor scenes, the networks designed for indoor scenario usually have smaller receptive radius to focus on object details, while the outdoor networks emphasize more on the object relations. 
Hence, the network specifically designed for indoor scenes usually performs bad on outdoor point cloud.
To deploy a EQ-Paradigm model on a particular application scenario, we need to carefully choose a practical embedding stage network for extracting support features. To enable a universal structure for all the 3D scenes, like the ResNet~\cite{ResNet} and ViT~\cite{VIT} in 2D, more exploration on the embedding stage network design is needed.

{\small
\bibliographystyle{ieee_fullname}
\bibliography{egbib}
}

\end{document}